%% file: main.tex
\documentclass[10pt,twocolumn,letterpaper]{article}
\PassOptionsToPackage{capitalize}{cleveref}

 \usepackage{cvpr}              
 

\definecolor{cvprblue}{rgb}{0.21,0.49,0.74}
\usepackage[pagebackref,breaklinks,colorlinks,allcolors=cvprblue]{hyperref}
\usepackage{algorithm}
\usepackage{graphicx}
\usepackage{amsmath}
\usepackage{bbm}
\usepackage{natbib}
\usepackage{algpseudocode}
\usepackage{cleveref}
\usepackage{multirow}
\usepackage{graphicx}
\usepackage{lscape}
\usepackage{xcolor}
\usepackage{amsmath}

\def\paperID{*****} 
\def\confName{CVPR}
\def\confYear{2025}

\title{Privacy Preserving Ordinal-Meta Learning with VLMs for Fine-Grained Fruit Quality Prediction}

\author{Riddhi Jain, Manasi Patwardhan, Aayush Mishra, Parijat Deshpande, Beena Rai \\
TCS-Research\\
Pune, India\\
{\tt\small \{jain.riddhi1, manasi.patwardhan, mishra.aayush1, parijat.deshpande, beena.rai\}@tcs.com}
}

\begin{document}
\maketitle
\input{sec/0_abstract}    
\input{sec/1_intro}
\input{sec/2_relatedwork}
\input{sec/3_dataset}
\input{sec/4_methods}
\input{sec/5_experiments}
\input{sec/6_results}

\input{sec/7_conclusion}
{
    \small
    \newpage
    \bibliographystyle{ieeenat_fullname}
    \bibliography{ref}
}


\end{document}

%% file: sec/0_abstract.tex
\begin{abstract}
To effectively manage the wastage of perishable fruits, it is crucial to accurately predict their freshness or shelf life using non-invasive methods that rely on visual data. In this regard, deep learning techniques can offer a viable solution. However, obtaining fine-grained fruit freshness labels from experts is costly, leading to a scarcity of data. Closed proprietary Vision Language Models (VLMs), such as Gemini, have demonstrated strong performance in fruit freshness detection task in both zero-shot and few-shot settings. Nonetheless, food retail organizations are unable to utilize these proprietary models due to concerns related to data privacy, while existing open-source VLMs yield sub-optimal performance for the task. Fine-tuning these open-source models with limited data fails to achieve the performance levels of proprietary models. In this work, we introduce a Model-Agnostic Ordinal Meta-Learning (MAOML) algorithm, designed to train smaller VLMs. This approach utilizes meta-learning to address data sparsity and leverages label ordinality, thereby achieving state-of-the-art performance in the fruit freshness classification task under both zero-shot and few-shot settings. Our method achieves an industry-standard accuracy of 92.71\%, averaged across all fruits.
\end{abstract}

%% file: sec/1_intro.tex
\section{Introduction}
\label{sec:formatting}

Given the limited shelf life of fruits, it is essential to monitor their journey from production to consumption in real-time to help reduce food waste \cite{r2,waste}. One compelling approach to monitoring the freshness of fruits within the food supply chain is through the use of deep learning. However, to be effective, it requires a lot of labeled data, in terms of  fruit images annotated with a freshness index \cite{r3,ml}. This is infeasible considering such fine-granular expert annotations being costly.

Large pre-trained Vision-Language Models (VLMs), like Gemini \cite{Gemini}, have demonstrated excellent performance on a range of multi-modal tasks, with their capacity to learn from context with no or very few demonstrations \cite{r6,goh2024vision,li2025benchmark}. This makes them a viable approach for handling scarce data. However, because of their proprietary nature and possible risk of disclosing sensitive data \cite{kibriya2024privacy}, organisations can be hesitant to employ them due to privacy concerns \cite{privacy,acquisti2016economics,jain2016big}. This problem is particularly significant in sectors where maintaining the secrecy of data is essential.

In the food industry, maintaining privacy is of utmost importance \cite{popper2007traceability,gregorczuk2022retail}. It prioritises protecting sensitive consumer data by discouraging uploads on the servers of the large proprietary models and thus, prefer on-site models \cite{foodprivacy}. Open-source VLMs such as Qwen2-VL \cite{qwen}, and blip \cite{blip}, can serve this purpose. However, with lesser parameters majority of these VLMs perform worse in zero-shot and few-shot scenarios as compared to large proprietary models \cite{raghunathan2005open}. They frequently fall short of their larger, proprietary equivalents in terms of accuracy and generalization capabilities.  Thus, there is need for a technique which can leverage these smaller VLMs as on-site models to  preserve data privacy, while at the same time maintaining performance comparable to larger VLMs. This would facilitate the food sector to apply efficient AI solutions striking a balance between  performance and data protection \cite{fp1,r9}.

\begin{figure*}
    \centering
    \includegraphics[width=\textwidth]{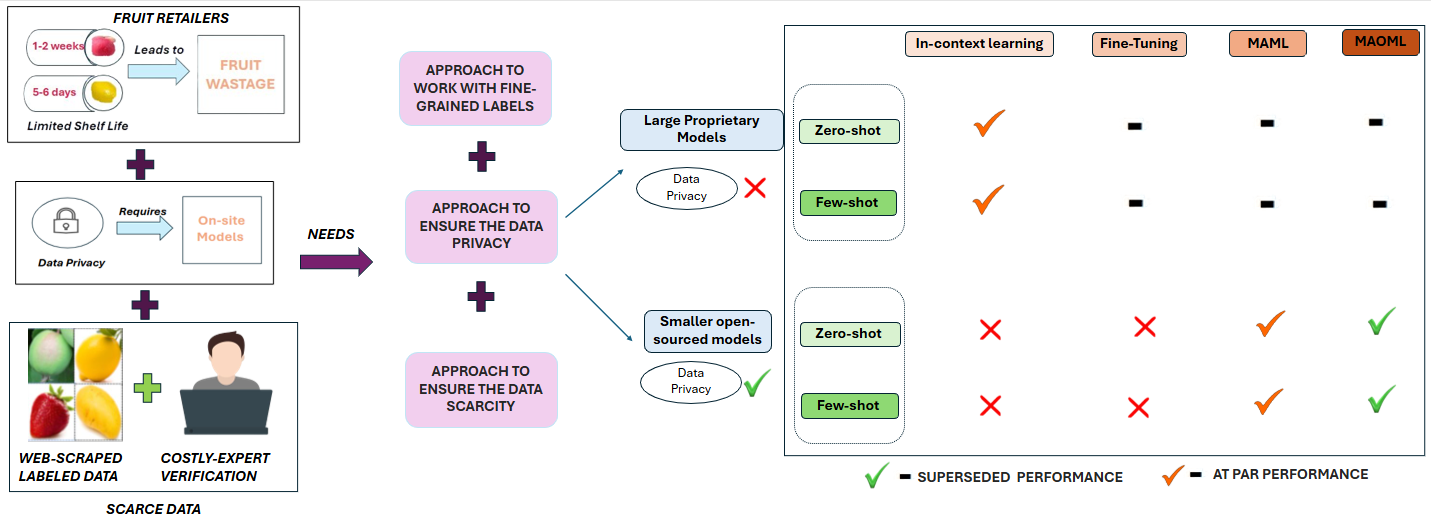}  
    \caption{Illustration of the need of an on-site model for fine-grained fruit freshness classification without compromising on the performance achieved by Large proprietary models. Our baselines and expected outcomes by our approach of Model Agnostic Ordinal Meta-Learning (MAOML).}
    \label{fig:my_label}
\end{figure*}

To address this demand, we first conduct experiments with smaller Vision Language Models (VLMs), fine-tuning them using the available limited data. Previous approaches \cite{r4,r5} meta-train Convolutional Neural Networks (CNNs) to mitigate the scarcity of labeled data by capturing common degradation patterns across different fruits. However, these approaches do not account for fine-grained labels. Given the frequent degradation of fruits \cite{degradation}, there is a need to detect subtle variations in their freshness index. Therefore, we incorporate more fine-grained labels, but we observe that meta-trained CNNs perform poorly under these conditions.

To further address this performance gap,  we leverage  the ordering in the fine-grained classes of the fruit quality, by augmenting meta-training ordinal regression loss \cite{ordinal}. This allows us to close the performance gap between smaller  VLMs and larger VLMs in in zero-shot as well as few-shot setting. This results in a viable solution to the challenge faced by food retail industry, addressing both high accuracy and data privacy needs \cite{retail}.

The main contributions of this work are:
\begin{itemize}
      \item We leverage scarce labelled data across distinct fruits to meta-train open-source small VLMs  for the fine-grained fruit quality predictions to achieve comparable performance with large VLMs in zero-shot setting and at the same time address privacy concerns.
      \item We define and apply Model-Agnostic Ordinal Meta-Learning (MAOML) technique which leverages ordinality in the fruit freshness labels to meta-train open-source small VLMs having 100x lesser number of parameters. The small VLMs trained with  MAOML supersedes the performance of a large VLM, in zero-shot and few-shot setting. 
\end{itemize}

%% file: sec/2_relatedwork.tex
\section{Related work}
\label{related_work}

\subsection{Fruit  freshness detection}
\cite{rizzo} has extensively reviewed various food freshness detection systems that uses different machine learning and deep learning methods, from which they conclude that such problems often suffer due to unavailability or lack of data. \cite{dutta2022transfer} use transfer learning and pairwise comparison within a Siamese network to predict age of various types of fruits and estimate their shelf life. \cite{fruit-f,r9,cnn} fine-tune a CNN model for more coarse granular labels (fresh and rotten) for fruit freshness classification.  \cite{mlforprediction} use pre-trained Vision Transformers for apple defect detection and banana ripeness estimation. \cite{ng2022fruit} adopts the meta-learning where a base network learns to extract meta-features and adapts to new types of fruits using only a few training samples. However, none of these approaches handle fine-grained fruit freshness annotations, which is taken into consideration as the core industrial setting in this work.

\subsection{Meta-Learning  Vision Language Models}
Meta-training VLMs have been explored in the literature for variety of tasks. \citep{r1} explores MAML meta-train VLMs for vision-language cross-lingual transfer. \cite{metamapper,meta-paper} meta-trains VLMs for visual question answering task by creating a  meta-mapper, which enables the training  without altering the frozen parameters of the model, making the training process more efficient as well as rapidly  adaptable to bind novel visual concepts to words by observing only a limited set of labeled examples. 
In our work we meta-train VLMs for fine-grained classification of fruit freshness by leveraging the ordinality in labels. We choose quantized parameter efficient method QLoRA\cite{qlora}  for the meta-training, which facilitates preserving most of the parameters of the original model to achieve generalizability across fruits with the scarce data, as well as makes the training more efficient and feasible in limited amount of available compute.

%% file: sec/3_dataset.tex
\section{Dataset}
\label{sec:dataset}

 We curate a dataset consisting of 10 different types of fruits listed in Table \ref{fruit-wise}. To capture a wide range of freshness quality stages, we defined five distinct quality classes for each fruit, viz. `Unripe' and `Early ripe', `Ripe', `Overripe' and `Bad'. We web-scrape images for each class of each fruit, by ensuring that we select only those images which are available under Creative Common licences. We provide an image and its label to a food scientist to validate the correctness of the label for the given fruit. With this expert verification, we take into consideration total 10 images per class per fruit (50 annotated images per fruit). We select 4 images per class of each fruit randomly to be included in the training set and remaining 6 fruits per class of each fruit are included in the test set (total 300 images). We have ensured that there is no overlap between the images of train and test set, thus there is no data leakage. We have two settings: 
 \begin{itemize}
     \item \textbf{Zero-shot:} We use 4 images per class for \(n - 1\) (9 in our case)  fruits  (total 180 images) for training and choose the test images of the n$^{th}$ (10$^{th}$ in our case) unseen fruit for inference. This setting simulates the real-life scenario, where an existing model trained on some set-of fruits is used to infer on newly introduced fruit with no availability of annotated data. 
     \item \textbf{Few-shot:}  We  select 4 images per class from all the n (10 in our case) fruits to form the training set of 200 images and perform inference on 300 test images of all the fruits. This setting simulates the real-life scenario, where there is availability of small amount of annotated data for all fruits to train a model. 
 \end{itemize} 

%% file: sec/4_methods.tex
\section{Methods}

\subsection{In-context Learning}
We assess pre-trained small as well as large vision-language models (VLMs) on the fruit ripeness classification task in  the zero-shot as well as the few-shot setting.  We task the model to categorize a fruit into one of the  ripeness  classes provided its image. We pose the classification as a generation task, by restricting the generation output to one of the class labels. Specifically,  we use the following  prompt: \textit{``You are a food expert specialized in identification of freshness of fruits. Classify the given fruit image into one of the freshness labels: 'Unripe,' 'Early ripe,' 'Ripe,' 'Overripe,' 'Bad.'"}. The model is expected to generate only one of the labels as the output. In zero-shot setting, we perform inference on the test sets of all the fruits to compute the average accuracy over all the fruits.

In the few-shot setting, we augment the prompt with the 4 images per class for each fruit (total 20 images), which are part of the training data of that fruit as discussed in section \ref{sec:dataset}, along with their labels. We perform inference on samples in the respective test set of those fruits to compute average accuracy over all the fruits.
We expect the few-shot approach to yield better performance than the zero-shot due to the availability of fruit-specific  demonstrations provided in the context.

\subsection{Fine-tuning} \label{sec:fine-tuning}
We perform task specific fine-tuning of the pre-trained small VLMs with our training set using the QLoRA framework leading to more efficient and feasible approach with the available compute and memory. In zero-shot setting, we perform fine-tuning with 9 fruits with  the  training set images as discussed in Section \ref{sec:dataset} and perform inference on the test set images of the 10$^{th}$ unseen fruit, leading to zero-shot inference on that fruit. We repeat this process, fine-tuning distinct models with the training sets of distinct 9 fruits, performing inference on the test images of the remaining fruit  and average the results over the test sets of all the fruits. Thus,  overall the inference is performed on 300 test images (6 fruits per class* 5 classes *10 fruits). Whereas, in few-shot setting we fine-tune a single model with the  200 images as discussed in Section \ref{sec:dataset} and perform inference on the test set images (300) of all the fruits with the same model. We expect the fine-tuning to result into better performance than in-context learning. However, we do not expect  large improvements in the performance due to the scarcity of the data, which is not sufficient to  train models with higher number parameters.

\subsection{Meta-Learning}
On the similar lines of \cite{r9}, use Model-Agnostic Meta-Learning (MAML) which is a meta-learning algorithm designed to enable quick adaptation to new tasks with limited data by learning an initialization that facilitates rapid parameter updates\cite{maml}. Our study treats various types of fruits as distinct tasks, where fine-grained quality labels represent classes, thus framing our objective as a multi-task multi-class classification problem. MAML optimizes model parameters in a way that minimizes the number of gradient steps needed for adaptation to novel tasks. MAML considers a model, denoted by f, such that f : x → y, where x represents the input fruit images and y is the corresponding fine-grained quality label. The model is trained to be able to adapt to a large number of fruits.

We perform MAML training of the models in QLora setting, which is feasible for us with the available compute. In zeros-shot setting, we meta-train with the training samples of 9 fruits and test it on the test samples of the 10$^{th}$ unseen fruit. In this setting, with the small number of labeled samples available, we expect MAML to learn the common degradation properties across the fruits in the train set, and efficiently  adapt to the unseen fruit. In few-shot setting, we meta-train a model with the train sets of all fruits and test it on the test-set of all the fruits, where now for each fruit we have 4 samples per class in the training set, leading to 20-shot setting.

\subsection{Model Agnostic Ordinal-Meta Learning (MAOML)}
Our task of fine-grained fruit quality classification has an inherent ordering
in its labels. A fruit would be considered ‘ripe’ only after it has passed the stages of ‘unripe’ and ‘early ripe’. Conventional classification losses, such as the multi-category cross-entropy, to such problems, are sub-optimal since they ignore the intrinsic order among the ordinal targets. Moreover, unlike in metric regression, we cannot quantify the distance between the ordinal ranks \cite{loss} as these ordinal categories have a predefined order but do not have a numerical distance defined between them.

Leveraging this ordinality of labels, as one of our main contribution, we use ordinal regression method that extends rank prediction into multiple binary label classification sub-tasks \cite{loss}. This approach involves three key steps: first, the expansion of rank labels into binary vectors; second, the training of binary classifiers on these extended labels; and finally, the computation of predicted rank labels based on the outputs of these binary classifiers. \\
In our supervised learning dataset \( D = \{ \mathbf{x}^{[i]}, y^{[i]} \}_{i=1}^{N} \), \( \mathbf{x}^{[i]} \in \mathcal{X} \) represents the fruit images for the \( i \)-th example out of \( N \) training examples, and \( y^{[i]} \) denotes its corresponding class label signifying the fine-grained quality. In an ordinal regression setting, we refer to \( y^{[i]} \) as the rank, which takes values from the ordered set \( \mathcal{Y} = \{ r_{1}, r_{2}, \ldots , r_{K} \} \), where \( r_{K} \succ r_{K-1} \succ \ldots \succ r_{1} \) and in our case, the ranks are the fine-grained quality levels such that \( \text{rotten} \succ \text{overripe} \succ \text{ripe} \succ \text{early ripe} \succ \text{unripe} \). The goal of ordinal regression is to learn a mapping \( h: \mathcal{X} \rightarrow \mathcal{Y} \) that minimizes a loss function \( L(h) \).

\cite{CORN} introduced CORN (Conditional Ordinal Regression for Neural Networks), a framework for rank-consistent ordinal regression. This approach employs the chain rule of conditional probabilities for predicting consistent ordinal ranks. Given a training set \( D = \{ \mathbf{x}^{[i]}, y^{[i]} \}_{i=1}^{N} \), CORN extends the ordinal rank labels \( y^{[i]} \) to binary labels \( y_{k}^{[i]} \in \{0,1\} \), indicating whether \( y^{[i]} \) exceeds rank \( r_{k} \). The objective is to predict whether a fruit image has passed a given quality level or not. The model sets up \( K-1 \) learning tasks corresponding to ranks \( r_{1}, r_{2}, \ldots , r_{K} \) in the output layer \cite{CORN}. CORN estimates conditional probabilities using subsets conditioned on the ranks, where \( f_{k}(\mathbf{x}^{[i]}) \) from the \( k \)-th binary task denotes the conditional probability.

\begin{equation}
f_{k}(\mathbf{x}^{[i]}) = \hat{P}\left( y^{[i]} > r_{k} \mid y^{[i]} > r_{k-1} \right) \tag{1}
\end{equation}

where the events are nested as: \( y^{[i]} > r_k \subseteq y^{[i]} > r_{k-1}\). The transformed, unconditional probabilities can then be computed by applying the chain rule for probabilities to the model outputs which also guarantees rank consistency among the \( K-1 \) binary tasks:

\begin{equation}
\hat{P}\left( y^{[i]} > r_{k} \right) = \prod_{j=1}^{k} f_{j}\left( \mathbf{x}^{[i]} \right) \tag{2}
\end{equation}

To train a CORN neural network using back propagation, we minimize the following loss function:

\begin{align}
L(X,y) = 
& -\frac{1}{\sum_{j=1}^{K-1} |S_j|} 
\sum_{j=1}^{K-1} \sum_{i=1}^{|S_j|} 
\Bigg[
    \log\!\big(f_j(\mathbf{x}^{[i]})\big)
    \mathbbm{1}\{ y^{[i]} > r_j \}
    \notag \\[4pt]
&\quad + 
    \big(1 - \log\!\big(f_j(\mathbf{x}^{[i]})\big)\big)
    \mathbbm{1}\{ y^{[i]} \leq r_j \}
\Bigg] 
\tag{3}
\end{align}

where  $|S_j|$ denote the size of the j-th conditional training set.

To utilize the inherent ordering of class labels while requiring minimal training data and to develop a model which works for multiple types of fruits, we combine the meta learning algorithm MAML with CORN and utilize the CORN loss function instead of cross-entropy loss. We also convert categorical labels to ordinal levels and modify the output layer to have c - 1 neurons for c classes as the objective is to predict the rank. The resulting Model-Agnostic Ordinal Meta Learning (MAOML) algorithm is provided in Algorithm \ref{alg:maoml}.

\begin{algorithm}[ht]
\caption{Model Agnostic Ordinal Meta Learning (MAOML)}
\label{alg:maoml}
\textbf{Require:} \( T = [T_1, \ldots, T_n] \): \( n \) tasks, each task \( T_i \) has \( c \) classes. \\
\textbf{Require:} \( \alpha, \beta \): step size hyper-parameters
\begin{algorithmic}[1]
\State Update the output layer to have \( c - 1 \) neurons
\State Convert the class labels to ordinal levels using \( \hat{y} = [1] \cdot y + [0] \cdot (c - 1 - y) \)
\State randomly initialize \( \theta \)
\While{not done}
    \For{all \( T_i \in T \)}
        \State Sample \( K \) data points \( D = \{ x^{(j)}, y^{(j)} \} \) from \( T_i \)
        \State Evaluate \( \nabla_{\theta} L_{T_i} (f_{\theta}) \) using \( D \) and \( L_{T_i} \) in Eq. 3 (CORN Loss)
        \State Compute adapted parameters with gradient descent:
        \State \( \theta'_i = \theta - \alpha \nabla_{\theta} L_{T_i} (f_{\theta}) \)
        \State Sample data points \( D'_i = \{ x^{(j)}, y^{(j)} \} \) from \( T_i \) for the meta-update
    \EndFor
    \State Update \( \theta \leftarrow \theta - \beta \nabla_{\theta} \sum_{T_i \sim p(T)} L_{T_i} (f_{\theta'_i}) \) using each \( D'_i \) and \( L_{T_i} \) in Eq. 3
\EndWhile
\end{algorithmic}
\end{algorithm}

%% file: sec/5_experiments.tex
\section{Experiments}

We use following small VLMs for our experiment: (i) Qwen2-VL-2B-Instruct\footnote{\hyperlink{https://huggingface.co/Qwen/Qwen2-VL-2B-Instruct}{https://huggingface.co/Qwen/Qwen2-VL-2B-Instruct}} \cite{qwen}, (ii) Qwen2-VL-7B-Instruct\footnote{\hyperlink{https://huggingface.co/Qwen/Qwen2-VL-7B-Instruct}{https://huggingface.co/Qwen/Qwen2-VL-7B-Instruct}} \cite{qwen}, (iii) blip-flan-t5-xl\footnote{\hyperlink{https://huggingface.co/Salesforce/blip2-flan-t5-xl}{https://huggingface.co/Salesforce/blip2-flan-t5-xl}}\cite{blip} with 3.94B parameters.. We use Gemini\cite{Gemini} as the large VLM. We use resnet-50\footnote{\hyperlink{https://huggingface.co/microsoft/resnet-50}{https://huggingface.co/microsoft/resnet-50}} \cite{resnet} and resnet-152 \footnote{\hyperlink{https://huggingface.co/microsoft/resnet-152}{https://huggingface.co/microsoft/resnet-152}} \cite{rajasree2022application} as our baseline models to benchmark our results against \cite{r9}. We also tested for smaller VLMs like llava\cite{llava} but the results were not up to the mark.

During the zero-shot inference using VLMs, we set the temperature to 0.1 for less variance in the output with a max\_length set to 50. For fine-tuning the ResNet-50 models, we use a learning rate of 0.00007, and we train with a batch size of 150 for 14 epochs. For all the VLM fine-tuning we use a learning rate of 0.0002 and we train with a batch size of 3  for epochs ranging from 7 to 9 with image size of 100x100. We meta-train the ResNet-50 model using MAML and MAOML, with a learning rate of 0.0007, batch size of 250 for  5 epochs and image size of 100x100. For VLMs in both the settings, we train the models  with a batch size of 2 for  8 epochs, with the learning rate of 0.0006 and image size of 100x100. For all the experiments,  hyper-parameters are tuned using GridSearch\footnote{\hyperlink{https://www.dremio.com/wiki/grid-search/}{https://www.dremio.com/wiki/grid-search/}}. We conduct all experiments of QLoRA training of VLMs on V100 CUDA GPU with 32 GB RAM. For training of Qwen-VL-2B-Instruct (without QLoRA) we use, batch size of 2 for 10 epochs, with the learning rate of  0.0007 and image size of 100x100. We conduct all experiments of this setting on A100 CUDA GPU with 40 GB RAM.
Treating food freshness detection as multi-class classification task, we evaluate using classification accuracy as the metric.

    
    \begin{table*}[ht]
    \centering
    \caption{Average (\%) accuracy  across test sets of all the fruits. IC: In-Context Learning, FT: Fine-tuning, \underline{\textbf{Bold and Underline: Best Performance}}, \textbf{Bold: Surpassing Gemini Performance}}
    \label{final_results_table}
    \small
    \resizebox{\textwidth}{!}{
    \begin{tabular}{|l|c|c|c|c|c|c|c|c|}
        \hline
         & \multicolumn{4}{c|}{\textbf{Zero-Shot}} & \multicolumn{4}{c|}{\textbf{Few-Shot}} \\
        \hline
         & \textbf{IC} & \textbf{FT} & \textbf{MAML} & \textbf{MAOML} & \textbf{IC} & \textbf{FT} & \textbf{MAML} & \textbf{MAOML} \\
        \hline
        ResNet-50\cite{resnet} & - & 25.31 & 56.90 & 61.20 & - & 28.90 & 59.27 & 65.49 \\
        \hline
        Resnet-152 & - & 29.49 & 62.13 & 69.92 & - & 29.81 & 67.25 & 74.42\\
        \hline \hline
        Blip-flan-t5-xl(3.94B) (QLoRA)\cite{blip} & 33.43 & 49.39 & 68.12 & 75.37 & 36.04 & 50.36 & 70.83 & 78.15 \\
        \hline
        Qwen2-VL-2B-Instruct (QLoRA)\cite{qwen}  & 33.20 & 54.05 & 81.34 & \textbf{84.98} & 34.52 & 63.58 & 84.81 & \textbf{89.22}\\
        \hline
        Qwen2-VL-2B-Instruct \cite{qwen}  & 38.52 & 55.20 & 79.25 & 85.99 & 38.82 & 71.29 & \textbf{88.35} & \textbf{92.13}\\
        \hline
        Qwen2-VL-7B-Instruct (QLoRA)\cite{qwen} & 43.04 & 56.84 & \textbf{85.96} & \underline{\textbf{90.28}} & 43.04 & 59.13 & \textbf{88.48} & \underline{\textbf{92.71}} \\
        \hline \hline
        Gemini\cite{Gemini} & \textbf{83.33} & - & - & - & \textbf{84.92} & - & - & - \\
        \hline
    \end{tabular}
    }
    \label{Results Table New}
\end{table*}

%% file: sec/6_results.tex
\section{Results and Discussion}
With our experiments we try to address the following research questions:

\textbf{RQ1: Do Vision-Language Models (VLMs) contribute towards improving the accuracy of fruit freshness prediction task, especially when compared to conventional CNN based pre-trained image classification models?} \\
With an emphasis on their ability to generalize across different datasets without task-specific knowledge, this inquiry seeks to investigate the benefits of employing VLMs in fruit freshness prediction task, over traditional CNN based pre-trained image classification models. 
As demonstrated in Table \ref{Results Table New}, the experiments showcase that VLMs,   perform exceptionally well on fruit freshness classification task, treated as class label generation task, for all three training mechanisms, viz. Fine-tuning, MAML and MAOML. We find even open-sourced VLMs surpass the baseline ResNet-50 model with a significant margin in both zero-shot and few-shot settings. The in-context learning setting with all VLMs, without any prior exposure to task specific data, yields better performance than fine-tuned ResNet-50. Among open-sourced VLMs, Qwen-VL-7B-Instruct performs the best. There is a huge gap between the performance of ResNet trained with our method (MAOML) and in-context learning  performance of the proprietary VLM, Gemini. Moreover, there is still a performance gap between the in-context learning  of the smaller open-source VLMs and Gemini. As illustrated in Table \ref{fruit-wise}, this is mainly because the open-source models perform very poorly on most of the fruits except banana, probably because of the lack of pre-training data for those fruits, whereas Gemini yield good performance across most of the fruits except apple and pineapple.

\begin{table*}[ht]
    \centering
    \small
    \caption{Fruit-wise \% accuracy for Qwen2-VL-7B-Instruct and Gemini}
    \label{fruit-wise}

    \begin{tabular}{|l|c|c|c|c|c|c|c|c|c|c|c|}
    \hline
    
         & \multicolumn{5}{c|}{Zero-Shot} & \multicolumn{5}{c|}{Few-Shot}\\
        \hline
        Fruits & \multicolumn{4}{c|}{Qwen2-VL-7B-Instruct} & Gemini & \multicolumn{4}{c|}{Qwen2-VL-7B-Instruct }& Gemini\\
        \hline
         & IC & FT& MAML & MAOML & IC &  IC & FT & MAML & MAOML& IC\\
        \hline
        Pineapple & 25.00 &43.92 & 78.96 & \textbf{86.62} & 65.50 & 25.00 & 44.68 & 80.30 & \underline{\textbf{87.92}} & 65.53 \\
        Guava & 29.17 & 70.13 & 83.99 & \textbf{90.28} & 79.13 & 29.17 & 70.83 & 85.38 & \underline{\textbf{91.02}} & 79.13  \\
        Pear & 35.24 & 48.17 & 98.10 & \textbf{98.59} & 89.35 & 35.24 & 48.89 & 98.47 & \underline{\textbf{98.92}} & 89.60
        \\
        Apple & 55.56 & 79.01 & 79.03 & \textbf{86.81} & 62.96 & 55.56 & 80.00 & 80.78 & \underline{\textbf{88.20}} & 63.36
        \\
        Pomegranate & 37.50 & 44.86 & 75.45 & \textbf{88.16} & 83.49 & 37.50 & 45.00 & 76.12 & \underline{\textbf{89.47}} & 83.49
        \\
        Mango & 40.43 & 54.43 & 85.65 & \textbf{86.73} & 81.13 & 40.43 & 61.70 & 95.16 & \underline{\textbf{96.74}} & 85.03
        \\
        Lemon & 42.67 & 37.52 & 92.72 & 95.00& \textbf{95.10} & 42.67 & 37.78 & 93.89 & \underline{\textbf{95.21}} & 95.12
        \\
        Strawberry & 45.45 & 48.70 & 84.47 & 87.66 & \textbf{89.10} &45.45&48.89& 85.20 & \underline{\textbf{90.20}} & 89.13
        \\
        Papaya & 47.50 & 68.63 & 81.18 & 83.18& \textbf{87.55} & 47.50 & 80.05 & 89.45 & 89.65& \underline{\textbf{91.75}}
        \\
        Banana & 71.87 & 73.06 & \underline{\textbf{100.00}} & \underline{\textbf{100.00}}& \underline{\textbf{100.00}} & 71.87 & 73.51 & \underline{\textbf{100.00}} & \underline{\textbf{100.00}} & \underline{\textbf{100.00}}
        \\
        \hline
    \textbf{Average} & 43.04 & 56.84 & 85.96 & \textbf{90.28} & 83.33 & 43.04 & 59.13 & 88.48 & \underline{\textbf{92.71}} & 84.92 \\
        
        \hline
    \end{tabular}
    \end{table*}

\textbf{RQ2: Does few-shot setting add value for the fruit freshness prediction task?}\\
In industrial setting, with a new addition of a fruit, soliciting fine-grained fruit freshness annotations for only few images can be viable. These annotations can facilitate in few-shot in-context learning setting as well as to re-train a model with addition of annotated images for the new fruit, leading to improvement in the performance of the novel fruit. This inquiry seeks to explore if such few-shot annotations for an unseen fruit can be useful. As illustrated in Table \ref{Results Table New}, for larger models, such as Qwen2-VL-7B-Instruct and Gemini few-shots help in achieving only marginal improvement in the performance with the in-context learning set-up, with improvements in only few fruits such as Mango and Papaya (Table \ref{fruit-wise}).   Whereas, comparatively smaller models such as Qwen2-VL-2B-Instruct and Blip-flan-t5-xl shows larger improvements with few-shot in-context learning over zero-shot. This shows that the larger models may have reached its saturation as far as in-context learning is concerned for fruit freshness detection task. With all the models we observe substantial improvements with few-shots, when trained with distinct training mechanisms, demonstrating the advantage of inclusion of few-shots of the new fruit for training.
    
\textbf{RQ3: Does fine-tuning smaller open-source VLMs facilitate fruit freshness prediction?} \\
This inquiry seeks to explore how fast and accurately the smaller VLMs  can adapt to fruit freshness classification task, by fine-tuning with the scarcely available data. 
As observed in Table \ref{Results Table New} fine-tuning of smaller VLMs gives a significant boost in their performance over in-context learning.
It is observed that smaller the VLM , larger is the increase in its average accuracy. A similar trend is recorded in the Table \ref{fine} where it is observed that the smallest VLM  with the least number of parameters showed the highest accuracy spike upon fine-tuning. This is mainly because of the scarcity in the data making it less feasible to tune model with more parameters \cite{compare}.  With Table \ref{fruit-wise} we can observe that fine-tuning can yield substantial improvements only for certain fruits such as Guava, Apple, Papaya.
Thus, as expected, smaller VLMs fine-tuned with the available scarce data perform better than zero-shot or few-shot in-context learning. However, the results of these fine-tuned models are still inferior as compared to  zero-shot performance of larger VLM (Gemini).
    
    \begin{table}[ht]
    \centering
    \caption{\% increase in the performance of the VLMs in few-shot setting with distinct training mechanisms over in-context learning. FT:  fine-tuning }
    \label{fine}
    \small
    \begin{tabular}{|l|c|c|c|}
        \hline
        Models & FT &  MAML&  MAOML  \\
          
   \hline
        
        Qwen2-VL-7B-Instruct & 37.38 & 106.27 & 107.29\\
        \hline
        Blip-flan-t5-xl (3.94B)& 39.73 & 96.53 & 116.84\\
         \hline
        Qwen2-VL-2B-Instruct & \textbf{84.18} & \textbf{145.68} & \textbf{158.46}\\
        \hline
    \end{tabular}
    \end{table}   

\textbf{RQ4: Does meta-training of open-source VLMs facilitate fruit freshness prediction?} \\
Meta-learning is one of the best methods to deal with scarce data\cite{ml}. Hence, with the scarce nature of our dataset, this inquiry seeks to explore if meta-training of smaller open-source VLMs performs better than fine-tuning.
As illustrated in Table \ref{Results Table New}, we observe a significant improvement in overall performance after MAML training of VLMs. In few-shot setting, Qwen2-VL-7B-Instruct model with MAML training surpasses the results of Gemini, helping us to achieve our goal. However, in zero-shot setting the performance is comparable. With training of all the parameters (without QLoRA),  even a much smaller model Qwen2-VL-2B-Instruct model surpass the performance of Gemini, with MAML training in few-shot setting but not in zero-shot setting.  However, it requires much higher compute.  
On the similar lines of fine-tuning based approach, as observed in Table \ref{fine}, the \% improvement in performance of MAML over in-context learning is higher for smaller models. However, as opposed to fine-tuning based approach, MAML training offers higher performance with larger Qwen2-VL model, consistently across all the fruits (Table \ref{fruit-wise}). This is mainly because of the capability of meta-learning approach to deal with scarce data with a higher capacity model.

    \begin{table}
    \centering
    \caption{Average accuracy(\%) of  Qwen2-VL-7B-Instruct on distinct labels across all the fruits for in few-shot setting.}
    \label{labels}
    \small 
    \begin{tabular}{|c|c|c|c|c|}
        \hline
        \textbf{Label} & \textbf{IC} & \textbf{FT} & \textbf{MAML} & \textbf{MAOML}\\
        \hline
        Unripe & 40.04 & 64.88 & 95.85 & 95.89 \\
        Early Ripe & 37.21 & 39.21 & 49.38 &  70.37\\
        Ripe & 51.41 & 71.02 & 100.00 & 100.00 \\
        Overripe & 39.17 & 57.61 & 100.00 & 100.00 \\
        Bad & 47.42 & 62.93 &97.17 &  97.29\\
        \hline
        \textbf{Average} & 43.04 & 59.13  & 88.48  & 92.71  \\
        \hline
    \end{tabular}
    \end{table}    
\textbf{RQ5: Does ordinal-regression loss adds value for the fruit freshness prediction task?} \\
With combined effect of meta-learning and ordinal regression loss, exploiting the inherent ordering of fruit quality classes, this inquiry seeks to explore if MAOML facilitates us to surpass the performance of the proprietary models. As observed in Table \ref{final_results_table},  with our approach of  MAOML training offers substantial improvements in results for all models in both zero-shot and few-shot settings. More importantly, both Qwen2-VL-2B-Instruct and Qwen2-VL-7B-Instruct models surpass the performance of Gemini, in both zero-shot as well as few-shot settings.  In the few-shot setting, Qwen2-VL-2B-Instruct leads to an increase of \textbf{5.06\%}, while Qwen2-VL-7B-Instruct leads to an increase of \textbf{6.31\%} in the zero-shot and \textbf{9.17\%} in the few-shot setting. Best performing Qwen2-VL-7B-Instruct surpasses the Gemini in-context learning performance for all the fruits except Papaya in few-shot learning setting, whereas Lemon, Strawberry and Papaya in zero-shot setting  (Table \ref{fruit-wise}). Thus, MAOML trained Qwen2-VL-7B-Instruct and Gemini can act as complementary models for distinct fruits.  In few-shot setting, QLoRA MAOML training of  Qwen2-VL-7B-Instruct allows us to achieve similar performance of MAML trained (without QLoRA) Qwen2-VL-2B-Instruct model, but with much lesser compute requirement for training. Whereas in zero-shot setting, QLoRA training yield comparable results for unseen fruits, preserving  generalization capability of the base-model. 
With Table \ref{labels}, for fine-tuning and MAML setting, we observe a non-uniform improvement in label-wise  accuracy, over in-context Learning, with our best performing open-source model (Qwen2-VL-7B-Instruct), with   very less improvement for `Early Ripe' fruits. Whereas, with MAOML we observe uniform improvement in the performance across for all the labels including `Early Ripe', leading to overall improvement in the average accuracy. This is mainly due to capability of MAOML to learn the ordinality in labels.

%% file: sec/7_conclusion.tex
\section{Conclusion}
With an assumption of data scarcity, fine-granular labels and the need of privacy preserving on-site models, in this paper, we have combined the meta-learning and ordinal regression strategies to develop an algorithm, Model Agnostic Ordinal Meta Learning (MAOML) to train ope-source Vision-Language Models for fine-grained fruit quality classification. 
With this approach, Qwen2-VL-7B-Instruct yields us the state-of-the-Art industry-standard performance of \textbf{90.28\%}  and \textbf{92.71\%} accuracy in zero-shot and few-shot setting, surpassing the performance of a much larger proprietary model Gemini.